# EVALUATING THE ACCURACY OF CLASSIFICATION ALGORITHMS FOR DETECTING HEART DISEASE RISK

Alhaam Alariyibi[1], Mohamed El-Jarai[2] and Abdelsalam Maatuk[3]

[1, 2]Department of Computer Science, Benghazi University, Benghazi, Libya
[3]Department of Information System, Benghazi University, Benghazi, Libya

## ABSTRACT

*The healthcare industry generates enormous amounts of complex clinical data that make the prediction of disease detection a complicated process. In medical informatics, making effective and efficient decisions is very important. Data Mining (DM) techniques are mainly used to identify and extract hidden patterns and interesting knowledge to diagnose and predict diseases in medical datasets. Nowadays, heart disease is considered one of the most important problems in the healthcare field. Therefore, early diagnosis leads to a reduction in deaths. DM techniques have proven highly effective for predicting and diagnosing heart diseases. This work utilizes the classification algorithms with a medical dataset of heart disease; namely, J48, Random Forest, and Naïve Bayes to discover the accuracy of their performance. We also examine the impact of the feature selection method. A comparative and analysis study was performed to determine the best technique using Waikato Environment for Knowledge Analysis (Weka) software, version 3.8.6. The performance of the utilized algorithms was evaluated using standard metrics such as accuracy, sensitivity and specificity. The importance of using classification techniques for heart disease diagnosis has been highlighted. We also reduced the number of attributes in the dataset, which showed a significant improvement in prediction accuracy. The results indicate that the best algorithm for predicting heart disease was Random Forest with an accuracy of 99.24%.*

## KEYWORDS

*Data Mining, Heart Disease, Weka Tool, Classification Techniques.*

## 1. INTRODUCTION

Today, the healthcare industry stores huge amounts of healthcare data that contains a large of valuable facts and information. However, the access to the knowledge that the efficiency of the healthcare systems is lacking [1]. Therefore, it is important to discover such knowledge with techniques that have efficient performance with expected high rate of accuracy. Data Mining (DM) is a technology used to analyze large-scale data sets to obtain significant and useful results using advanced artificial intelligence techniques that performance accurately and efficiently [2, 3].

In many countries heart disease is one of the fatal diseases which causes maximum casualties. According to the WHO organization, heart diseases are a significant health concern as around17.9 million people died from CVD diseases in 2019, which represents 32% of all global deaths. 85% of these deaths were due to a heart attack and stroke [4]. On other hand, most heart diseases may be passive and not detected until people experience symols or symptoms of a heart attack, heart failure, or an arrhythmia. Some physicians may rely on some of the factors and signs that increase the risk of heart disease such as age, sex, chest pain, blood pressure, cholesterol, blood sugar, fasting, the maximum heart rate, and family history of heart disease [5,6]. However, it is difficult to manually determine the odds of getting heart disease based on risk factors [7]. In medicine, the diagnosis of diseases is an important task but still complicated and intricate job that needs to be executed accurately and efficiently. Regrettably, diagnosis that is based only on the knowledge and experience of the physician may lead to undesirable results. Hence, misdiagnosis or delayed diagnosis can have serious consequences for patients, including reduced treatment efficacy and poorer outcomes, while diagnostic methods can be invasive, expensive, and in some cases inaccessible. Therefore, it is important to detect heart disease as early as possible so that it can begin to be treated, addition to following up on medical advices and medications [4]. Machine-learning techniques have shown a promise in accurately predicting and diagnosing various diseases including heart disease. Hence, computer-aided diagnosis systems can enable early diagnosis and potentially increase survival rates while reducing the need for expert analysis of predicted diseases. In healthcare, predictive DM techniques are more commonly used [8].

Based on the above, there is a need to develop and evaluate a predictive model that can accurately predict heart diseases by analyzing medical datasets. Such a model can effectively avoid and treat data-related problems such as missing, incorrect, and inconsistent data collection [9]. The model can be beneficial to the medical sector that directly assist physicians to obtain accurate diagnoses of heart disease and thus make intelligent clinical decisions efficiently in less time. Hence, this paper aims to exploit a predictive DM technique called classification in a structure referred to as a model to improve the accuracy of heart disease diagnosis. In the classification stage, three classifiers called J48 decision tree, Naïve Bayes and Random Forest are used to predict and diagnose heart disease. Furthermore, the proposed model optimizes the classification model performance by utilizing a feature selection approach. A local Libyan medical dataset was used to train and test the proposed model. The performance of the utilized model was evaluated and the results were analyzed using various metrics. Moreover, a comparative study has been conducted to improve the performance in terms of accuracy, sensitivity and specificity.

This paper is organized as follows: Section II is dedicated to presenting some of the research studies concerned with the diagnosis of heart disease using predictive DM techniques. Section III provides a brief description of the classification process for the proposed model. Section IV describes the experimental setup and results, and the discussion of the results in presented in Section V. Section VI concludes the paper.

## 2. RELATED WORK

Over the past two decades, a lot of research has been performed on diagnosing heart disease using different DM techniques and algorithms.

Three AI-based methods, namely Decision Trees, Naïve Bayes and Neural Networks are applied for predicting cardiovascular or heart disease [2]. The dataset used is from the UCI repository that consisted of 14 attributes with 668 records. The obtained results show that the Decision Tree gave the highest accuracy with 98.54% compared to Naïve Bayes with 85.01% and Neural Networks with 81.83%.

Another three techniques, i.e., Naïve Bayes, J48 Decision Tree and Bagging were used to predict heart disease [6]. Different experiments were conducted using the Weka tool. The obtained results were compared using 10 cross-validations that were used to measure the unbiased estimate of these prediction algorithms. The dataset, which was taken from the Hungarian Institute of Cardiology, has 76 raw attributes, but only 11 attributes were selected for the experiments. The results show that the Bagging has the highest accuracy, i.e., 85.03 % and J48 Decision Tree and Naïve Bayes have 84.35% and 82.31% accuracy respectively.

Boshra Baharami et al. [10], have evaluated different classification techniques such as the J48 Decision Tree, k-Nearest Neighbours (k-NN), Naïve Bayes, and Sequential Minimal Optimization (SMO). The dataset used contains 209 records and 8 features that are collected from a hospital in Iran. On the other hand, the selection technique is used to extract the important features. The Weka tool is used for implementing the classification algorithms. Some performance evaluation measures such as accuracy, precision, sensitivity, specificity, F-measure, and area under the ROC curve are used to evaluate and compare the performance evaluation techniques. In addition, a 10-fold cross-validation technique is used to test the mining techniques. The experimental results showed that the J48 Decision Tree is the best classifier for heart disease diagnosis with an accuracy of 83.732%.

The tool Weka was used to implement the algorithms J48 Decision Tree, Naïve Bayes and Multi-Layer Perceptron Neural Network (MLPNN) in a study described in [11] The data were collected from the Ibb Hospital, Yemen. The experimental results are evaluated by metrics like accuracy, TP-rate, Precision, F-Measure and ROC graph. The outcome of predictive classification techniques on the same dataset reveals that all classification techniques used with 3 selected attributes outperformed the same techniques with all attributes. However, the most effective classifier to predict patients with heart disease was a J48 Decision Tree classifier with a classification accuracy of 100%.

A study that focused on the heart data by applying various DM techniques through the Weka tool is described in [12]. Three algorithms were used for predicting heart diseases; namely Decision Tree, Neural Network and Naïve Bayes. The dataset was used from the UCI repository. The results were compared using the metrics of accuracy, Precision, and F-measure, ROC curve value, TP rate and FP rate. The results showed that Naïve Bayes classification algorithms have the highest accuracy among all that is 82.914%.

Some DM techniques, i.e., Naïve Bayes and Decision Tree have been applied to perform an analysis on heart disease diagnosis [13]. Different experiments were conducted using the Weka 3.6.0 tool. The dataset was taken from the UCI Machine Learning Repository, which consists of 294 records with 13 attributes. The results concluded that the Naïve Bayes outperformed the Decision tree techniques, which obtained an accuracy of 85.03%.

Three different DM techniques are used to predict cardiovascular heart disease, namely Naïve Bayes classifiers, KNN, and SVM [14]. The machine learning algorithms are implemented using R programming language. The algorithms were evaluated based on the classification of accuracy and Confusion Matrix table. The dataset was taken from the UCI Machine Learning Repository with 14 attributes. The results showed that the Naïve Bayes held the highest accuracy rate with 86.6% in diagnosing the presence of heart disease.

A hybrid model is proposed based on the Decision Tree technique, i.e., the C4.5 algorithm, that is combined it with ANN [15]. The dataset was taken from the UCI repository, where the accuracy, sensitivity, and specificity of the individual classifier and hybrid technique are used to compare algorithms. The hybrid DT model was analysed and compared with the C4.5 algorithm and ANN on the same data set, and proved to be more accurate with an accuracy of 78.14%

A study was conducted and developed an automated system that aimed at reducing the tests conducted by patients, as well as saving cost and time for both the analyst and the patients [16]. The study considered Naïve Bayes, Decision Tree and Neural Network for the experimentation.

Based on the experimental result, it is observed that the neural network with 15 attributes outperformed the Decision tree and the Naïve Bayes with an accuracy score of 100% against 99.62 and 90.74% obtained from the Decision Tree and Naïve Bayes respectively.

Most of the previous work have been conducted on data sets that were collected and obtained from different sources and with certain specifications. However, we used a recent sample obtained from the Benghazi Medical Center (BMC dataset) in Libya. In this paper, we focused on implementing an effective and accurate diagnosis of heart disease using classification algorithms on the BMC dataset. Moreover, the feature selection method was applied by reducing the number of attributes in the dataset to increase the efficiency of the classification algorithms and the accuracy of heart disease prediction.

## 3. DATASET

The dataset used in this study was obtained from Benghazi Medical Center and was collected in the period between the years 2018 to 2022. The dataset contains 529 instances with 14 medical attributes, including class labels. There are 8 categorical attributes and 6 numeric attributes. The target attribute was identified as the predictable attribute with a value of 1 for patients with heart disease and 0 for non-infected patients. The description of the dataset is given in Table 1.

Table 1. Attributes of heart disease dataset.

| Sr. No | Attribute | Description | Value | Type |
|---|---|---|---|---|
| 1 | Sex | Male or female | 1 for male<br>0 for female | Nominal |
| 2 | Age | Age in years | Continuous | Numeric |
| 3 | Cp | Chest pain type | 1 = typical type<br>2 = typical type agina<br>3 = non-agina pain<br>4 = asymptomatic | Nominal |
| 4 | Trestbps | Resting blood pressure | Continuous value in mm hg | Numeric |
| 5 | Chol | Serum cholesterol | Continuous value in mm / dl | Numeric |
| 6 | Fbs | Fasting blood sugar | 1 > 120 mg/dl<br>0 < 120 mg/dl | Nominal |
| 7 | Restecg | Resting electrographic results | 0 = normal<br>1 = having ST_T wave abnormal<br>2 = left ventricular hypertrophy | Nominal |
| 8 | Thalach | Maximum heart rate achieved | Continuous value | Numeric |
| 9 | Exang | Exercise included agina | 0 = no<br>1 = yes | Nominal |
| 10 | Old peak | ST depression induced by exercise relative to rest | Continuous value | Numeric |
| 11 | Slope | The slope of the peak exercise ST segment | 1 = unsloping<br>2 = flat<br>3 = down-sloping | Nominal |
| 12 | Ca | Number of major vessels colored by floursopy | 0-3 value | Numeric |
| 13 | Thal | Defect type | 3= normal<br>6 = fixed<br>7 = reversible defect | Nominal |
| 14 | Target | Diagnosis of heart disease | 1 = infected<br>0 = not infected | Nominal |

## 4. PROPOSED MODEL FRAMEWORK

The model framework proposed in this study consists of five stages, as shown in Figure 1. Firstly, the dataset is pre-processed using a range of techniques to enhance and prepare the data appropriately. Next, the feature selection is applied to the pre-processed data to increase its variability and robustness. Thirdly, the dataset is split and fed into the model, utilizing the classification techniques to optimize the model performance. Once the model is trained, it is evaluated on the data to obtain classifications. The following sections describe the framework phases.

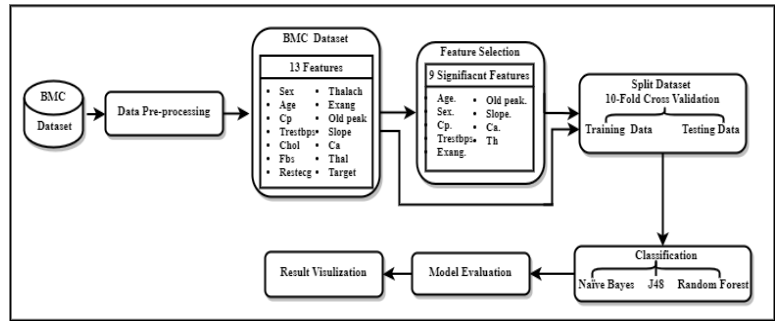

Figure 1. The framework of the proposed model.

### 4.1. Data Pre-processing

The Data pre-processing phase concerns transforming unclean data into a pure dataset. This leads to increase the efficacy and accuracy of a machine learning model. pre-processing was applied, although some problems such as missing data, inconsistent data, and duplicate data were not addressed. In the pre-processing phase, the process of data transformation into the formal needed is involved. In our dataset, all the nominal attributes were transformed into numeric data. The dataset was obtained as an Excel file, which is then converted to the Attribute-Relation File Format (ARFF) as the Weka tool supports the ARFF file.

### 4.2. Feature Selection

Feature selection is a data reduction technique that is applied to the dataset in order to identify the most significant features of heart diseases. Moreover, reducing the size of the dataset by removing irrelevant features helps to construct an accurate model and enhance the learning performance [17]. For our experiments, the selection feature was applied using the filter and search method of the Weka tool. In this regard, we have used a supervised attribute filter called CfsSubsetEval that maintains or improves the performance of learning algorithms. It is widely used for medical diagnosis due to its effectiveness to solve diagnostic and prediction problems. It performs concerning the accuracy of learning schemes using the reduced feature sets. It can evaluate the best subset of attributes in the given dataset by considering the individual predictive ability of each feature as well as the degree of redundancy between them. Subsets of features with high intercorrelation are preferred, while exhibiting low intercorrelation [18]. The Best First searching method was then applied to add the best of the original remaining attributes to the dataset after each iteration. The method selects 9 attributes from the total of 13 attributes, without including the predicted attribute. These selected attributes are Age, Sex, Cp, Trestbps, Exang, Old Peak, Slope, Ca and Thal.

### 4.3. Classification

The proposed model is for the diagnosis of heart disease using classification algorithms using a common tool in DM called Weka version 3.8.6. In the classification phase, a total of three classification algorithms namely, J48, Random Forest, and Naïve Bayes are used to classify the existing datasets according to the important features. Many researchers are applied and developed different DM techniques, but most of them indicate that how these classifiers can effectively be used in the medical field respect to the individual algorithms [7]. Before classifications, we considered the datasets in two parts, i.e., training data and testing data. Separately, each classifier was trained twice using the training dataset with all attributes and with the selection ones. The performance of these classifiers was tested with the test dataset to evaluate their performance efficiency.

Naive Bayes one of the most classifiers used for machine learning analysis due to its simplicity to implement. It is widely used for medical diagnosis because it has effective ability to solve diagnostic and prediction problems [19]. The Naïve Bayes classifier is proposed based on the Bayes theorem. Naïve Bayes is a probability-based classifier assuming that all features are conditionally independent, where changes in one feature do not affect another feature [7]. Bayesian classifiers satisfy the general principle of the Bayesian method, in which the minimum error rate for classification is obtained, and the maximum efficiency is achieved, allowing for no Bayesian methods to be used for classification [20].

Many researches indicate that Random Forest classifier is a flexible and easy to use that effectively predicates a great result most of the time. It is one of the most used algorithms; due to its simplicity and diversity [19]. It is based on tree technology, in which a forest with many trees is created. It is an ensemble algorithm that combines multiple algorithms, in which a set of decision trees is built from a random sample of the training set [7]. After repeating the process with several random samples, the final decision is taken based on the majority vote. The Random Forest algorithm can handle missing values effectively, but it may run into overshoots that can be avoided by applying appropriate parameter tuning.

The J48 algorithm plays an important role in the field of medical diagnosis which is quite fast, popular and the output is easily interpretable. It is one of the best decision tree algorithms that shown useful accuracy in the diagnosis of heart disease. It can handle continuous data in numerical forms. This algorithm can be managed with pruning algorithms that simplify classification rules without reducing prediction accuracy. It performs error-based pruning where only the most important features are kept, whereby they lower the error rates. The J48 treated the data as input and generated a decision tree as output. In this algorithm, univariate trees are created as output. The classification rules are constructed in the form of a decision tree [7, 21].

### 4.4. The Performance Evaluation

The performance of the proposed model was assessed using several evaluation techniques, which include accuracy, specificity, and sensitivity. Hence, these measures are calculated from the outcomes of the Confusion Matrix, which displays the frequency of correct and incorrect predictions. It is used to evaluate the performance of the classification algorithm by associating the actual target values for the response variable, heart patients, with the trained model's prediction [19]. The accuracy is the percentage of correct predictions relative to the total number of cases evaluated. The sensitivity is mathematically defined as the ratio of the total number of True-Positive patients to the sum of the number of True-Positive and False-Negative patients. It measures the percentage of correctly classified patients who have heart disease. Similarly, the specificity is the ratio of the total number of True Negatives to the sum of the number of True

Machine Learning and Applications: An International Journal (MLAIJ) Vol.10, No.4, December 2023Negatives and False Positive patients. It measures the percentage of correctly classified patients who have no heart disease. The equations 1, 2 and 3 describe how they are calculated, where TP stands for True Positive, TN stands for True Negative, FP stands for False Positive, and FN stands for False Negative.

$$\text{Accuracy} = \frac{T_P + T_N}{T_P + F_P + T_N + F_N} * 100 \quad (1)$$

$$\text{Sensitivity} = \frac{T_P}{T_P + F_N} \quad (2)$$

$$\text{Specificity} = \frac{T_N}{T_N + F_P} \quad (3)$$

## 5. EXPERIMENTS AND RESULTS

We conducted three experiments, where the proposed model was trained using Weka software. The Weka offers novice users a tool to identify hidden information from database and file systems with easy-to-use options and visual interfaces [22]. The BMC dataset used in these experiments consists of 529 records in the heart disease database. For the experimental purpose, the evaluation of classification performance models was evaluated by using the 10-fold cross-validation method. In this approach, the entire dataset is randomly sampled into ten subsets, nine of which are utilized as testing data and the remaining subset as training. After being processed ten times, the results are obtained by calculating each ten iterations on average [7]. We opted for this method as it increases the validation of classification and prevents random and invalid results. The DM classification algorithms namely J48, Random Forest, and Naïve Bayes are implemented using the Weka tool. The three experiments were conducted with and without feature selection to assess the effect of feature selection. All the classifiers were evaluated based on the classification metrics including accuracy, sensitivity, and specificity that were calculated by the Confusion Matrix. The performance of the proposed model is demonstrated in all three experiments.

### 5.1. Experiment I

In Experiment I, we evaluated the performance of the J48 classifier in predicting heart disease. First, we tested the model with the BMC dataset with all 14 attributes and without removing any attribute. Then, we implement the classifier on all dataset instances with selected attributes. The result of a Confusion Matrix for both situations is shown in Table 2.

Table 2. Confusion Matrix for experiment I.

| Classifier | | Predicted 1 | Predicted 0 |
|---|---|---|---|
| J48 with all attributes | Actual 1 | 298 | 3 |
| | Actual 0 | 10 | 218 |
| J48 with selected attributes | Actual 1 | 298 | 3 |
| | Actual 0 | 9 | 219 |

By evaluating the model on the selected dataset, we found that the model achieved an accuracy of 97.54% and 97.73% without and with using Feature Selection, respectively as indicated in Table 3 with the results of the rest of the metrics.

Table 3. The output performance for experiment I.

| Classifier | Accuracy | Specificity | Sensitivity |
|---|---|---|---|
| J48 with all attributes | 97.54% | 95.61% | 99% |
| J48 with selected attributes | 97.73% | 96.05% | 99% |

## 5.2. Experiment II

Experiment II was conducted to assess the effectiveness of the Naïve Bayes classifier while executed. For Naïve Bayes with all attributes, 280 and 182 instances are correctly classified into target classes, but 67 records are incorrectly classified. The Confusion Matrix for this experiment is illustrated in Table 4.

Table 4. Confusion Matrix for experiment II.

| Classifier | | Predicted 1 | Predicted 0 |
|---|---|---|---|
| Naïve Bayes with all attributes | Actual 1 | 280 | 21 |
| | Actual 0 | 46 | 182 |
| Naïve Bayes with selected attributes | Actual 1 | 277 | 24 |
| | Actual 0 | 35 | 193 |

Using feature selection, there was an increase in the value of accuracy and sensitivity metrics while the value of specificity was described. The classification accuracy of using the Select Feature (88.85%) is higher than the classification accuracy of all attributes (87.33%) as shown in Table 5.

Table 5. The output performance for experiment II.

| Classifier | Accuracy | Specificity | Sensitivity |
|---|---|---|---|
| Naïve Bayes with all attributes | 87.33% | 79.82% | 93.02% |
| Naïve Bayes with selected attributes | 88.85% | 84.65% | 92.02% |

## 5.3. Experiment III

In Experiment III, predicting heart disease was conducted using the Random Forest classifier. Table 6 shows the Confusion Matrix, which represents the outcomes of this experiment.

Table 6. Confusion Matrix for experiment III.

| Classifier | | Predicted 1 | Predicted 0 |
|---|---|---|---|
| Random Forest with all attributes | Actual 1 | 300 | 1 |
| | Actual 0 | 4 | 224 |
| Random Forest with selected attributes | Actual 1 | 301 | 0 |
| | Actual 0 | 4 | 224 |

In Experiment III, the model achieved an optimal accuracy of 99.24% using the Feature Selection. All attributes, 300 and 224 records are correctly classified, but a total of 5 records are incorrectly classified. The performance measures for this experiment are illustrated in Table 7.

Table 7. The output performance for experiment III.

| Classifier | Accuracy | Specificity | Sensitivity |
|---|---|---|---|
| Random Forest with all attributes | 99.05% | 98.25% | 99.67% |
| Random Forest with selected attributes | 99.24% | 98.25% | 100% |

## 6. DISCUSSION

In this study, we employed the binary classifier based on a supervised DM algorithm for classification to predict heart diseases. Different performance measures are adopted to evaluate the proposed model including accuracy, sensitivity, and specificity. The summary of the classifiers' performance in the proposed model is shown in Table 8.

Table 8. The summary of classifiers performance.

|   | Classifier | Accuracy | Specificity | Sensitivity |
|---|---|---|---|---|
| With all attributes | J48 | 97.54% | 95.61% | 99% |
|   | Naïve Bayes | 87.33% | 79.82% | 93.02% |
|   | Random Forest | 99.05% | 98.25% | 99.67% |
| With selected attributes | J48 | 97.73% | 96.05% | 99% |
|   | Naïve Bayes | 88.85% | 84.65% | 92.02% |
|   | Random Forest | 99.24% | 98.25% | 100% |

The proposed prediction model was built using three classification algorithms (i.e., J48, Naïve Bays, and Random Forest) in two different stages. In the initial stage, the model was performed without Feature Selection. In the second stage, the classifiers were performed with Feature Selection. In both stages, the data were split using the 10-across-validation method. In terms of accuracy analysis, Figure 2 indicates that the lowest accuracy is 87.33% and the highest is 99.24%. The accuracy was used to determine the ability of the model to predict diseases correctly. In both stages, with and without Feature Selection, the Random Forest model had the highest accuracy of 99.24% and 99.05%, respectively. The J48 accuracy comes next with 97.73% and 97.54%, respectively. Nonetheless, the Naïve Bayes had the lowest accuracy of 87.33% with all features, and with the Feature Selection was better, i.e., 88.85%. This shows that the RF algorithm is the best classifier to predict heart patients. These results of experiments indicate that there was an increase in the accuracy performance of classifiers of 0.19% for both Random Forest and J48 algorithms and 1.85% for Naïve Bayes through Future Selection. Figure 2 indicates the increase in accuracy of Random Forest, J48 and Naïve Bayes classifiers using Feature Selection.

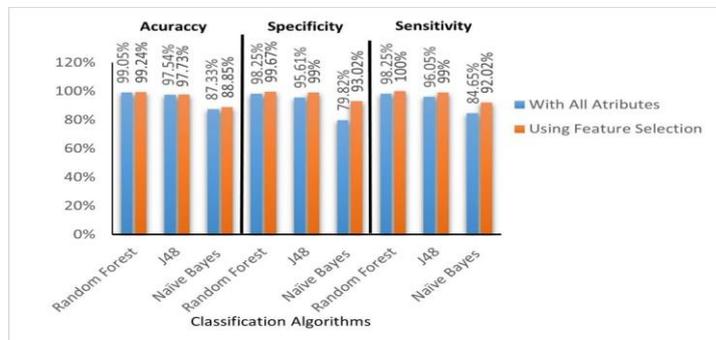

Figure 2: Comparison of accuracy, specificity, and sensitivity.

Other performance measures the specificity and sensitivity that are used to compare the results and achieve a remarkable performance and shows very close differences. The sensitivity was used to find the proportion of true patients suffering from the disease. On the other hand, the specificity was also used to determine the true proportion of true patients who are not affected by heart disease. Figure 3 presents a comparison of these performance measures in the classifiers. The specificity and sensitivity of Random Forest with all attributes were 98.25%% and 99.67%, respectively, and Feature Selection was 98.25% and 100%. Observably, the Random Forest algorithm with the Feature Selection correctly identifies all patients who have the disease. Similarly, the test correctly classified 98.25% of patients as not having heart disease, while 1.75% of them were misclassified.

As shown in Figure 2, the results show that using the Feature Selection is a good strategy for improving the specificity of Naïve Bayes and J48 classifiers and only the sensitivity of the Random Forest classifier. Nevertheless, the Naïve Bayes classifier was better with all attributes at classifying patients who were affected by heart disease than with using the Feature Selection. On the other hand, the specificity of the Random Forest in both stages has the same value (99.25%) and the sensitivity of the J48 classifier in both stages has the same value (99%). As shown in Figure 3, it was easier for the Random Forest classifier with selected attributes to identify negative cases correctly compared to the other classifiers. In contrast, using the Naïve Bayes classifier with all attributes struggled slightly to identify negative cases correctly compared to the other classifiers. One important thing observed was that all the models were doing better at predicting positive cases compared to negative ones.

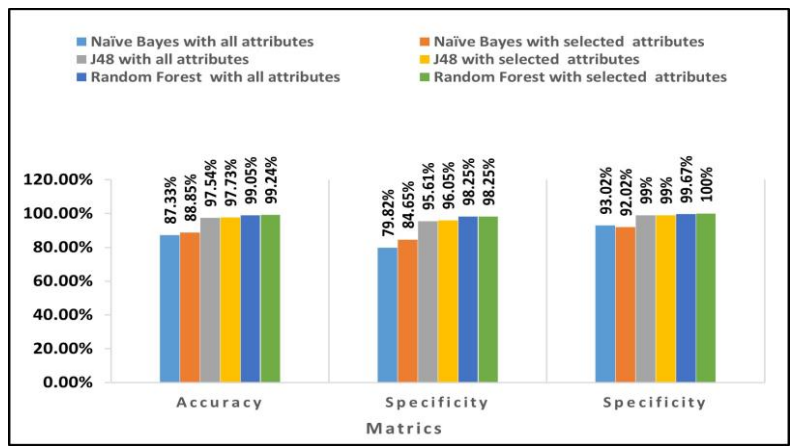

Figure 3. Comparison of model classifiers performance.

One interesting aspect that emerged from the analysis of results was that the performance was further enhanced using the Feature Selection method. The results illustrated that the accuracy of the implemented algorithms has increased by using the Feature Selection method. In contrast, experiment III yielded the most satisfactory results. The results found that, in both phases, the Random Forest performs the highest accuracy, sensitivity, and specificity compared to the other classifiers used in the proposed model.

## 7. CONCLUSION

This paper describes a study to examine the effect of utilizing predictive DM techniques on the accuracy of heart disease prediction. The study used a new curated dataset of heart diseases, obtained from Benghazi Medical Center, that contains a total of 529 instances with 14 medical attributes. To classify this dataset, we employed classification algorithms namely, J48, Random

Forest, and Naïve Bayes. The CfsSubsetEval filter and the Best First searching method were utilized to select significant features. Consequently, nine attributes are selected from all attributes. The performance of the proposed model was evaluated by standard metrics such as accuracy, sensitivity, and specificity. Three experiments were conducted with and without feature selection to study the effect of feature selection, and the Weka tool was used for the classification of the dataset. The results findings have proven the effectiveness and efficiency of classification algorithms for the diagnosis and prediction of heart diseases using datasets labelled efficiently. The results highlighted the highest improvement in accuracy yielded the feasibility of using the feature selection method to diagnose heart diseases. The Random Forest algorithm had the highest accuracy of prediction, sensitivity, and specificity, respectively with misclassified errors of 0.7561%. In addition, the results indicated that the Random Forest algorithm is the most appropriate for heart disease prediction with remarkable accuracy rates of 99.24%.

## ACKNOWLEDGEMENTS

The authors appreciate the contribution of Benghazi Medical Center as a source of the home dataset.

## REFERENCES

[1]   Patel, Jaymin and TejalUpadhyay, Dr and Patel, Samir, "heart disease prediction using machine learning and data mining technique," Heart Disease, vol. 7, no. 1, pp. 129--137, 2015.

[2]   P. Shetgaonkar and S. Aswale, "Heart Disease Prediction using Data Mining Techniques," International Journal of Engineering Research & Technology (IJERT), Vol. 10 Issue 02, Feb. 2021. [Online]. Available: www.ijert.org.

[3]   I. Yoo, P. Alafaireet, M. Marinov, K. Pena-Hernandez, R. Gopidi, J. F Chang and L. Hua, "Data mining in healthcare and biomedicine: a survey of the literature," Journal of medical systems, 36, pp. 2431-2448, May 2011.

[4]   "CardioVascular Diseases (CVDs)," 11 June 2021. [Online]. Available: https://www.who.int/news-room/fact-sheets/detail/cardiovascular-diseases-(cvds). [Accessed 23 July 2023].

[5]   P. Singh, S. Singh, and G. S. Pandi-Jain, "Effective heart disease prediction system using data mining techniques," Int J Nanomedicine, vol. 13, pp. 121–124, 2018.

[6]   V. Chaurasia and S. Pal "Data Mining Approach to Detect Heart Diseases," International Journal of Advanced Computer Science and Information Technology (IJACSIT), vol. 2, no. 4, pp. 56-66, 2013.

[7]   C. B. C. Latha and S. C. Jeeva, "Improving the accuracy of prediction of heart disease risk based on ensemble classification techniques," Inform Med Unlocked, vol. 16, July 2019.

[8]   N. Jothia, N. Abdul Rashidb and W. Husainc, "Data Mining in Healthcare – A Review," Procedia Computer Science, Dec. 2015.

[9]   R. El-Bialy, M. Salamay, O. Karam and M. Khalifa, "Feature Analysis of Coronary Artery Heart Disease Data Sets," Procedia Computer Science, pp. 459-468, 2015.

[10]  B. Bahrami and M. H. Shirvani, "Prediction and Diagnosis of Heart Disease by Data Mining Techniques," Journal of Multidisciplinary Engineering Science and Technology (JMEST), vol. 2, issue. 2, Feb. 2015. [Online]. Available: www.jmest.org.

[11]  B. Mohamad Al-Maqaleh and A. Mohamad Gasem Abdullah, "Intelligent Predictive System Using Classification Techniques for Heart Disease Diagnosis," International Journal of Computer Science Engineering (IJCSE), vol. 6, no. 06, June 2017.

[12]  U. Shafique, F. Majeed, H. Qaiser, and I. U. Mustafa, "Data Mining in Healthcare for Heart Diseases," International Journal of Innovation and Applied Studies, vol. 10 no. 4, March 2015. [Online]. Available: http://www.ijias.issr-journals.org/

[13]  M. R. Thansekhar and N. Balaji, "Heart Disease Diagnosis using Predictive Data Mining," in 2014 International Conference on Innovations in Engineering and Technology (ICIET'14), vol. 3, issue 3, March 2014 [Online]. Available: http://www.cs.waikato.ac.nz/ml/weka.

[14] S. Anitha and N. Sridevi, "HEART DISEASE PREDICTION USING DATA MINING TECHNIQUES," Journal of Analysis and Computation (JAC), vol. XIII, issue. II, Feb. 2019. [Online]. Available: www.ijaconline.com,

[15] S. Maji and S. Arora, "Decision Tree Algorithms for Prediction of Heart Disease," in Lecture Notes in Networks and Systems, pp. 447–454, 2019.

[16] N. Bhatla and K. Jyoti, "An Analysis of Heart Disease Prediction using Different Data Mining Techniques," International Journal of Engineering Research & Technology (IJERT), vol. 1 issue 8, Oct. 2012. [Online]. Available: www.ijert.org

[17] Kaushalya Dissanayake, Md Gapar Md Johar, "Comparative Study on Heart Disease Prediction Using Feature Selection Techniques on Classification Algorithms," Applied Computational Intelligence and Soft Computing, pp. 17, Nov. 2021.

[18] Hall Mark A. Correlation-based feature selection for machine learning. The University of Waikato; 1999.

[19] M. Murad Hossain, S. Khurshid, K. Fatema, M. Z. Hasan, and M. Kamal Hossain, "Analysis and Prediction of Heart Disease Using Machine Learning and Data Mining Techniques," Canadian Journal of Medicine, pp. 36-44, 2021.

[20] A. Methaila, P. Kansal, H. Arya, and P. Kumar, "Early Heart Disease Prediction Using Data Mining Techniques," Academy and Industry Research Collaboration Center (AIRCC, pp. 53–59), Aug. 2014.

[21] S. M. Alzahani, A. Althopity, A. Alghamdi, B. Alshehri, and S. Aljuaid, "An Overview of Data Mining Techniques Applied for Heart Disease Diagnosis and Prediction," Lecture Notes on Information Theory, vol. 2, no. 4, 2015.

[22] Kulkarni, E. G and Kulkarni, R. B, "Weka powerful tool in data mining," International Journal of Computer Applications, 2016.

| J48 with all attributes | | |
|---|---|---|
| **Statistic** | **Value** | **95% CI** |
| Sensitivity | 99.00% | 97.12% to 99.79% |
| Specificity | 95.61% | 92.08% to 97.88% |
| Accuracy (*) | 97.54% | 95.83% to 98.69% |

| J48 with selected attributes | | |
|---|---|---|
| **Statistic** | **Value** | **95% CI** |
| Sensitivity | 99.00% | 97.12% to 99.79% |
| Specificity | 96.05% | 92.64% to 98.18% |
| Accuracy (*) | 97.73% | 96.07% to 98.82% |

| Naïve Bayes with all attributes | | |
|---|---|---|
| **Statistic** | **Value** | **95% CI** |
| Sensitivity | 93.02% | 89.53% to 95.63% |
| Specificity | 79.82% | 74.02% to 84.83% |
| Accuracy (*) | 87.33% | 84.20% to 90.05% |

| Naïve Bayes with selected attributes | | |
|---|---|---|
| **Statistic** | **Value** | **95% CI** |
| Sensitivity | 92.03% | 88.37% to 94.82% |
| Specificity | 84.65% | 79.30% to 89.07% |
| Accuracy (*) | 88.85% | 85.85% to 91.40% |

| Random Forest with all attributes | | |
|---|---|---|
| **Statistic** | **Value** | **95% CI** |
| Sensitivity | 99.67% | 98.16% to 99.99% |
| Specificity | 98.25% | 95.57% to 99.52% |
| Accuracy | 99.05% | 97.81% to 99.69% |

| Random Forest with selected attributes | | |
|---|---|---|
| **Statistic** | **Value** | **95% CI** |
| Sensitivity | 100.00% | 98.78% to 100.00% |
| Specificity | 98.25% | 95.57% to 99.52% |
| Accuracy (*) | 99.24% | 98.08% to 99.79% |